\documentclass{article}

\usepackage[final,nonatbib]{neurips_2025_ml4ps}

\usepackage[utf8]{inputenc} % allow utf-8 input
\usepackage[T1]{fontenc}    % use 8-bit T1 fonts
\usepackage{hyperref}       % hyperlinks
\usepackage{url}            % simple URL typesetting
\usepackage{booktabs}       % professional-quality tables
\usepackage{amsfonts}       % blackboard math symbols
\usepackage{nicefrac}       % compact symbols for 1/2, etc.
\usepackage{microtype}      % microtypography
\usepackage{xcolor}         % colors
\usepackage[style=numeric-comp,sorting=none]{biblatex}

\usepackage{amsmath,amssymb}
\usepackage{braket}
\usepackage{dsfont}
\usepackage{subcaption}
\usepackage{svg}
\usepackage{cleveref}
\usepackage{todonotes}
\usepackage{csquotes}
\usepackage{amsthm}
\theoremstyle{plain}
\newtheorem{theorem}{Theorem}

\theoremstyle{definition}

\theoremstyle{remark}

\DeclareMathOperator*{\argmax}{arg\,max}

\renewcommand{\vec}[1]{\boldsymbol{#1}}
\newcommand{\mat}[1]{\boldsymbol{#1}}
\newcommand{\E}[1]{\mathbb E\left[#1\right]}

\newcommand{\X}{\textbf{\text{X}}}
\newcommand{\Y}{\textbf{\text{Y}}}

\newcommand{\Z}{\textbf{\text{Z}}}

\DeclareMathOperator{\tr}{tr}

\crefname{section}{Sec.}{Secs.}
\crefname{figure}{Fig.}{Figs.}
\crefname{equation}{}{}

\title{Quantum Boltzmann Machines for Sample-Efficient Reinforcement Learning}

\author{%
	Thore~Gerlach
	\\
	University of Bonn\\
	\texttt{tgerlac1@uni-bonn.de} \\
	\And
	Michael~Schenk \\
	CERN \\
	\texttt{michael.schenk@cern.ch} \\
	\And
	Verena~Kain \\
	CERN \\
	\texttt{verena.kain@cern.ch} \\
}

\begin{document}

\maketitle

\begin{abstract}
	We introduce theoretically grounded Continuous Semi-Quantum Boltzmann Machines (CSQBMs) that supports continuous-action reinforcement learning. By combining exponential-family priors over visible units with quantum Boltzmann distributions over hidden units, CSQBMs yield a hybrid quantum-classical model that reduces qubit requirements while retaining strong expressiveness. Crucially, gradients with respect to continuous variables can be computed analytically, enabling direct integration into Actor-Critic algorithms. Building on this, we propose a continuous $Q$-learning framework that replaces global maximization by efficient sampling from the CSQBM distribution, thereby overcoming instability issues in continuous control.
\end{abstract}

\section{Introduction}
\label{sec:introduction}

%In recent years, Reinforcement Learning~(RL) methods are increasingly investigated for deployment in complex real-world scenarios~\cite{kent2024using}.
%However, one often faces the challenge of restricted access
%to training data.
%For instance, this holds true for particle \emph{beam control} in high-energy physics experiments at the European Organisation for Nuclear Research~(CERN)~\cite{gatignon2018physics,adli2018acceleration,bartosik2022performance,montbarbon2019cern}.
%This task gets increasingly challenging due to the need for higher beam intensities and smaller beam sizes, as well as the sheer number of diverse experimental requirements.
%While physics models are used for many beam-control tasks, several systems still rely on manual tuning.
%These systems present complex non-linear continuous-control optimization problems, where \emph{sample efficiency} is a crucial requirement.
%This is due to the limitation of data-taking opportunities in terms of restricted beam times in accelerator operation.
%However, classical RL algorithms often struggle to meet the stringent efficiency and stability requirements.

In recent years, Reinforcement Learning~(RL) has been increasingly investigated for deployment in complex real-world scenarios~\cite{kent2024using}.
One major challenge is the restricted access to training data, as in the case of \emph{beam control} in high-energy physics experiments at CERN~\cite{gatignon2018physics,adli2018acceleration,bartosik2022performance,montbarbon2019cern}.
This task is further complicated by demands for higher beam intensities, smaller beam sizes, and diverse experimental requirements.
While physics models support many beam tasks, several systems still rely on manual tuning.
These systems often present complex non-linear continuous-control optimization problems where \emph{sample efficiency} is critical due to limited beam times in accelerator operation.
Classical RL algorithms, however, often fail to meet the stringent efficiency and stability requirements.

Energy-based models, and in particular Boltzmann Machines~(BMs)~\cite{ackley1985learning}, offer a principled probabilistic framework for approximating value functions in RL, while being sample-efficient~\cite{sallans2004reinforcement,salakhutdinov2009deep,grathwohl2019your}.
%These models consist of interconnected visible and hidden binary neurons, where the visible units encode the data and the hidden units lend the model its representational power.
Since BMs can approximate any probability distribution arbitrarily well, they are capable of capturing highly non-linear correlations in data.
At the same time, general BMs are notoriously difficult to train, as sampling from a Boltzmann distribution is NP-hard~\cite{barahona1982computational}.
%In practice, this bottleneck is circumvented by considering models based on Restricted BMs~(RBMs), which assume a bipartite structure of the underlying connectivity graph. 
%RBMs also have a large representational power~\cite{le2008representational}, but with the cost of potentially needing exponentially many hidden units.
%Samples can be obtained with Markov Chain Monte Carlo methods~\cite{hinton2002training}, but obtaining high-quality samples is still hard~\cite{long2010restricted}.
Markov Chain Monte Carlo methods allow sampling from Boltzmann machines~\cite{hinton2002training}, yet obtaining reliable high-quality samples continues to be a major challenge~\cite{long2010restricted}.

Quantum Computing~(QC)~\cite{nielsen2010quantum} offers a remedy: due to the phenomena of \emph{superposition}, \emph{entanglement} and the probabilistic nature of QC, it is used for efficiently sampling from distributions with exponentially-sized spaces and highly-correlated variable structures~\cite{piatkowski2024quantum}.
%Quantum Annealers~(QAs)~\cite{kadowaki1998quantum,liu2018adiabatic,dixit2021training} and, gate-based quantum devices capable of preparing Gibbs states~\cite{chen2023quantum,PhysRevA.110.012445,rouze2024optimal}, have been proposed as samplers for Boltzmann distributions.
Quantum Annealers (QAs)~\cite{kadowaki1998quantum,liu2018adiabatic,dixit2021training} and gate-based quantum devices designed to prepare Gibbs states~\cite{chen2023quantum,PhysRevA.110.012445,rouze2024optimal} have been proposed as quantum samplers for Boltzmann distributions.
Thus, QC could accelerate BM training and prediction, thereby enhancing their practical relevance.
Furthermore, expressiveness of BMs is increased by introducing quantum terms into their energy function, leading to Quantum BMs~(QBMs).
While gradient computation is intractable for arbitrary connectivity~\cite{amin2018quantum}, restricting the structure yields trainable models~\cite{anschuetz2019realizing,coopmans2024sample,demidik2025expressive}.
Classical visible units enable efficient gradient computation, yielding Semi-QBMs~(SQBMs) that surpass their classical counterparts in expressiveness~\cite{demidik2025expressive}.

$Q$-learning is among the most sample-efficient RL methods~\cite{watkins1992q}, but its applicability is limited to discrete action spaces.
Proof-of-concept studies have demonstrated that quantum annealers can improve sample efficiency in this setting by leveraging SQBMs~\cite{crawford2018reinforcement,levit2017free}.
Extending $Q$-learning to continuous actions is notoriously unstable, as it requires global maximization over a non-linear function approximator, an intractable problem.
To overcome this, a hybrid Actor-Critic~(AC) framework with an SQBM-based critic was proposed~\cite{schenk2024hybrid}.
While AC methods naturally support continuous actions, they continue to struggle with action constraints, training instability, and sample inefficiency.
Moreover, these approaches often lack solid theoretical grounding and rely on approximations~\cite{suzuki1976relationship,levit2017free}, which can substantially degrade performance.

This work is motivated by these challenges and the lack of a framework for continuous-valued visible units for SQBMs.
We address the limitations of energy-based learning together with the instability of continuous-action RL, thereby advancing sample-efficient control methods for complex physical systems such as those at CERN and beyond.
Our main contributions are of theoretical nature and are summarized as follows (see also~\cref{fig:overview}):
\begin{itemize}
	\item We propose the first theoretically sound SQBM formulation with continuous visible units, called continuous SQBM~(CSQBM), which largely reduces the number of required qubits. Our formulation naturally extends classical continuous-valued BMs,
	%	and can be employed in energy-based RL for handling continuous action spaces
	\item Regarding the limitations of AC algorithms, we present the first continuous Q-learning algorithm based on sampling from the hybrid quantum-classical probability distribution of a CSQBM for obtaining the best action, overcoming recent assumptions on the expressiveness of the chosen Q-function approximation while maintaining sample efficiency.
\end{itemize}
% First theoretically sound QBM with continuous visible units and quantum energy-based RL framework with handling continuous action spaces
% unifiying framework of clamped QBMs and SQBMs
% Propose new methods for continuous actions utilizing sampling for maximizing Q-value
% Experimental results???

\section{Related Work}
\label{sec:related_work}

Quantum RL~(QRL) comprises several distinct approaches~\cite{meyer2022survey}.
At the most fundamental level, fully quantum formulations and subroutine-based methods (e.g., amplitude amplification~\cite{Brassard2000QuantumAA}) promise asymptotic advantages but typically require fault-tolerant hardware.
In contrast, variational approaches~\cite{jerbi2021parametrized} are better suited to near-term devices, replacing classical function approximators with parameterized quantum circuits, though they face challenges such as barren plateaus and noise.
Finally, energy-based models, in particular quantum Boltzmann machines (QBMs), exploit quantum sampling for richer representations and are able to capture multi-modalities.
%Benchmarking~\cite{neumann2023quantum,kruse2025benchmarking,meyer2025benchmarking}

By employing the free energy~(FE) of a QBM to approximate the $Q$-value, prior works~\cite{kent2024using,crawford2018reinforcement,levit2017free,neumann2020multi} demonstrated improved sample efficiency compared to classical neural networks~(NNs) in prototypical environments with discrete action spaces.
An extension to continuous-valued environments was investigated in~\cite{schenk2024hybrid}, where an AC approach was applied to the AWAKE beam line at CERN.
This method represents continuous states and actions by encoding them into Bernoulli-distributed binary visible units, thereby compromising the theoretical soundness of the BM model.
This forces reliance on finite-difference approximations for gradient computation over the visible units, a procedure that is both computationally inefficient and prone to numerical instability.

Several extensions of Boltzmann machines to continuous visible domains have been developed to better accommodate real-valued data.
Prominent examples include Gaussian-Bernoulli models~\cite{cho2011improved,melchior2017gaussian} and more general formulations based on exponential-family distributions~\cite{welling2004exponential,li2018exponential}.
While these approaches enhance expressiveness, they suffer from training instabilities caused by sampling difficulties.
To address this, a continuous-valued QBM has been proposed~\cite{bangar2025continuous}, leveraging imaginary-time evolution on photonic quantum hardware for more efficient sampling.
Although promising, this method is inherently tied to that specific platform and does not readily extend to other QC paradigms.

In continuous environments, AC algorithms often achieve state-of-the-art performance, yet they suffer from persistent challenges such as handling action constraints, training instability, and, most critically, sample inefficiency.
By contrast, $Q$-learning~\cite{watkins1992q} is highly sample-efficient but restricted to low-dimensional discrete action spaces, since it requires global maximization over the input of an NN---an intractable task due to high non-linearity~\cite{katz2017reluplex}.
While reformulations of the maximization step have been investigated~\cite{ryucaql,burtea2024constrained}, their scalability remains uncertain.
Sampling-based methods can efficiently approximate local optima~\cite{kalashnikov2018scalable,simmons2019q,perakis2022optimizing}, but they risk unstable or divergent training.
Alternatively, analytic solutions for the global optimum are possible under strong structural assumptions on the NN~\cite{gu2016continuous,plaksin2022continuous,amos2017input}, though at the cost of severely limiting representational power.

Our approach addresses these limitations by introducing CSQBMs, which combine a theoretically sound hybrid quantum–classical distribution: an exponential-family prior over the visible units and a quantum Boltzmann distribution over the hidden units.
Gradients with respect to the visible units can be computed analytically, making the model directly applicable within AC frameworks.
In addition, we propose a continuous $Q$-learning algorithm based on CSQBMs that enables efficient sampling for global $Q$-value maximization.

\section{Background}
\label{sec:background}

For notational convenience, we denote vectors/matrices by bold lower/upper case letters and the identity matrix as $\mat I$.
%where its dimension should be clear from the context.
Further, $\tr\left[\cdot\right]$ denotes the trace of a matrix.

\subsection{Quantum Boltzmann Machines}
\label{sec:qbm}

A BM is a recurrent binary neural network~(NN) and consists of two types of neurons: visible units $\vec v\in \{-1,1\}^n$ and hidden (latent) units $\vec h\in\{-1,1\}^m$.
The visible units are observed and encode the data, while hidden units give the model its representational power.
Weighted connections between units define a quadratic energy function $E(\vec v,\vec h)$, characterizing a Boltzmann distribution $p(\vec v,\vec h)\propto e^{-E(\vec{v},\vec h)}$.
It has the capability of approximating every distribution arbitrarily well, making BMs useful for learning tasks.
Since drawing exact samples from this distribution is intractable~\cite{barahona1982computational}, one inevitably faces a trade-off between sample quality and computational efficiency.

QBMs are promising in overcoming the sampling limitation.
Instead of considering binary units, QBMs assume every unit to be represented by a quantum bit (qubit).
Instead of taking a definite value in $\{-1,1\}$, the state of a qubit is represented by a $2$-dimensional complex vector $\ket{\psi}\in\mathbb C^2$
\begin{align*}
	\ket{\psi}=a\ket{0}+b\ket{1},\ \ket{0}=(1,0)^{\top},\ \ket{1}=(0,1)^{\top},\ a,b\in\mathbb C,\ |a|^2+|b|^2=1\;.
\end{align*}
Even though a qubit can exist in a mixture/\emph{superposition} of the basis states $\ket{0}$ and $\ket{1}$ (corresponding to $-1$ and $1$), its exact state cannot be observed.
The only information retrieval possible is through \emph{measurement}, which leads the state $\ket{\psi}$ to collapse to $\ket{0}$ with probability $|a|^2$ and to $\ket{1}$ with probability $|b|^2$.
Considering $N$ qubits with states $\ket{\psi}_1,\dots, \ket{\psi}_N$, their joint quantum state $\ket{\psi}$ is implicitly exponentially large, that is $\ket{\psi}=\ket{\psi}_1\otimes\dots\otimes\ket{\psi}_N\in\mathbb C^{2^N}$.
Through the quantum mechanical phenomenon of \emph{entanglement}, single qubits become strongly correlated, creating an exponentially large state space that classical computers cannot efficiently simulate.
Combined with their probabilistic nature, a system of $N$ qubits allows the encoding of arbitrary discrete probability distributions over a $2^N$-dimensional space.
Through measurement of the quantum state, QC can thus be used for efficiently obtaining samples.
A more sophisticated introduction into QC is given in~\cite{nielsen2010quantum}.

The energy of a quantum system can be described by a \emph{Hamiltonian}, which is a Hermitian matrix $\mat H\in\mathbb C^{2^N\times 2^N}$ ($\mat H^{\dagger}=\mat H$)  with $N=n+m$.
The Hamiltonian of a QBM takes the form
%\begin{equation}
%	\small 
%	\mat H 
%%	&= \sum_{(\P,\Q)\in\mathcal P_{\vec v\vec h}}\sum_{i\in[n],j\in[m]}W_{ij}^{\vec v\vec h,\P}\P_i\otimes\Q_{n+j} \\
%	=\sum_{i}\mat{w}^{\vec v}_{i,\cdot}\mat H^{\vec v}_i\otimes \mathbf{I}_m+
%	\sum_{i,j}\mat{w}^{\vec v\vec v}_{i,j,\cdot}\mat H^{\vec v}_i\otimes \mat H^{\vec v}_j+
%	\sum_{i,j}\mat{w}^{\vec v\vec h}_{i,j,\cdot}\mat H^{\vec v}_i\otimes \mat H^{\vec h}_j+
%	\sum_{i,j}\mat{w}^{\vec h\vec h}_{i,j,\cdot}\mat H^{\vec h}_i\otimes \mat H^{\vec h}_j
%	\;,
%\end{equation}
\begin{align}
	\mat H&=\mat H^{v}+\mat H^{vv}+\mat H^{vh}+\mat H^{h}+\mat H^{hh}
	=\sum_{\mat P}\mat H^{v}_{\mat P}+\mat H^{h}_{\mat P}+\sum_{\mat Q}\mat H^{vv}_{\mat P\mat Q}+\mat H^{vh}_{\mat P\mat Q}+\mat H^{hh}_{\mat P\mat Q}\;,
	%	&\mat H^{\cdot}_{\mat P\mat Q}=\sum_{i,j}w_{ij}^{\cdot,\mat P\mat Q}\mat P_i \mat Q_j,\quad \mat H^{\cdot}_{\mat P}=\sum_iw_i^{\cdot,\mat P}\mat P_i
	\label{eq:qbm_hamiltonian}
\end{align}
%\begin{equation}
%	\mat H
%	%	&= \sum_{(\P,\Q)\in\mathcal P_{\vec v\vec h}}\sum_{i\in[n],j\in[m]}W_{ij}^{\vec v\vec h,\P}\P_i\otimes\Q_{n+j} \\
%	=\sum_{i}\mat{w}^{\vec v}_{i}\mat H^{\vec v}_i+
%	\sum_{i,j}\mat{w}^{\vec v\vec v}_{i,j}\mat H^{\vec v\vec v}_{ij}+
%	\sum_{i,k}\mat{w}^{\vec v\vec h}_{i,k}\mat H^{\vec v\vec h}_{ik}+
%	\sum_{k}\mat{w}^{\vec h}_{k}\mat H^{\vec h}_k+
%	\sum_{k,l}\mat{w}^{\vec h\vec h}_{k,l}\mat H^{\vec h\vec h}_{kl}
%	\;,
%	\label{eq:qbm_hamiltonian}
%\end{equation}
%with $\mat H_i^{\vec v},\mat H^{\vec h}_k\in \mathbb C^{3\times 2^{N}\times 2^{N}}$ and $\mat H_{ij}^{\vec v\vec v},\mat H_{ik}^{\vec v\vec h},\mat H_{kl}^{\vec h\vec h}\in \mathbb C^{9\times 2^{N}\times 2^{N}}$ describing $3$-tensors which contain Hamiltonians for the three different Pauli operators, $\mat{w}^{\vec v}_{i},\mat{w}^{\vec h}_{k}\in \mathbb R^{3}$ and $\mat{w}^{\vec v\vec v}_{ij},\mat{w}^{\vec v\vec h}_{ik},\mat{w}^{\vec h\vec h}_{kl}\in \mathbb R^{9}$ are weight vectors and $i,j\in\{1,\dots,n\}$, $k,l\in\{1,\dots,m\}$, $N=m+n$.
where $\mat H^{\cdot}_{\mat P}=\sum_iw_i^{\cdot,\mat P}\mat P_i$ encode the linear biases for the visible and hidden units and $\mat H^{\cdot}_{\mat P\mat Q}=\sum_{i,j}w_{ij}^{\cdot,\mat P\mat Q}\mat P_i \mat Q_j$ encode pairwise interactions.
$\mat P,\mat Q\in\{\X,\Y,\Z\}$ are Pauli matrices and $\mat P_i$ denotes applying operator $\mat P$ to qubit $i$.
Choosing $\mat P=\mat Q=\Z$ leads to a classical BM, while using more Pauli terms leads to more \enquote{quantumness}.
Encoding weights into the three different Pauli basis configuration results in a significantly larger amount of weights, making QBMs more expressive than classical BMs.
For more details, we refer the reader to~\cite{demidik2025expressive}.

The underlying quantum Boltzmann distribution is characterized by the Gibbs state $\rho = {e^{-\beta \mat H}}/{\mathcal Z}\in\mathbb C^{2^{N}\times 2^N}$, where $\beta>0$ is an inverse temperature and $\mathcal Z=\tr \left[e^{-\beta \mat H}\right]$.
%\begin{align}
%	p(\vec v,\vec h)=\tr\left[\rho\mat{\Delta}_{\vec v\vec h}\right],\quad \rho = \frac{e^{-\beta \mat H}}{\mathcal Z},\quad \mathcal Z=\tr \left[e^{-\beta \mat H}\right],\quad \mat{\Delta}_{\vec v\vec h}=\ket{\vec v}\bra{\vec v}\otimes \ket{\vec h}\bra{\vec h}\;,
%\end{align}
The diagonal of $\rho$ encodes the probability of all configurations and the marginal distribution of the visible units is obtained by 
\begin{align}
	p(\vec v)=\tr\left[\rho_{\vec v}\right]=\frac{\tr\left[e^{-\beta \mat H}\mat{\Delta}_{\vec v}\right]}{\mathcal Z}
	=\frac{e^{-\beta F_{\mat H}(\vec v)}}{\mathcal Z},
	\quad F_{\mat H}(\vec v)=-\frac{1}{\beta}\log\tr\left[e^{-\beta \mat H}\mat{\Delta}_{\vec v}\right]\;,
\end{align}
where $\mat{\Delta}_{\vec v}$ is a projection matrix onto the configuration $\vec v$ and $F_{\mat H}(\vec v)$ is called the free energy of $\vec v$.

Even though QBMs are powerful in theory, training them in practice is intractable, since gradients can not be computed analytically for arbitrary Hamiltonian structures.
In~\cite{demidik2025expressive}, it was shown that gradients of $\tr\left[e^{-\beta \mat H}\mat{\Delta}_{\vec v}\right]$ are analytically computable when only Pauli-$\Z$ terms are used for the visible units, since in that case $\mat H\mat{\Delta}_{\vec v}=\mat{\Delta}_{\vec v}\mat H$.
This leads to a reduced number of trainable weights, however, these models were shown to still be more expressive than their classical BM counterpart.

\begin{figure}[!t]
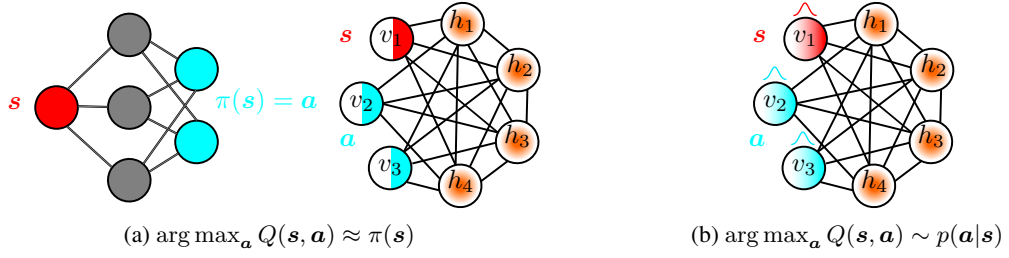

	\centering
	\begin{subfigure}{0.6\textwidth}
		\centering
		\includesvg[height=0.12\textheight]{img/actor_new.svg}
		\caption{$\argmax_{\vec a} Q(\vec s,\vec a)\approx\pi(\vec s)$}
	\end{subfigure}
	\hfill
	\begin{subfigure}{0.3\textwidth}
		\centering
		\includesvg[height=0.12\textheight]{img/csqbm_rotated.svg}
		\caption{$\argmax_{\vec a} Q(\vec s,\vec a)\sim p(\vec a|\vec s)$}
	\end{subfigure}
	\caption{Illustration of our proposed CSQBM and how it enables continuous-action $Q$-learning. We overcome the limitations of current AC approaches using SQBMs (a) by introducing theoretically sound continuous SQBMs~(CSQBMs), which utilize exponential-family priors (e.g. Gaussian) (b). The best action is obtained by sampling from the hybrid quantum–classical distribution.}
	\label{fig:overview}
\end{figure}

\subsection{Free Energy-Based Reinforcement Learning}
\label{sec:ferl}

For an in-depth description of RL and the underlying Markov Decision Processes~(MDPs), we refer the reader to~\cite{sutton1998reinforcement}.
Generally, RL is a framework for sequential decision-making in which an agent learns to maximize the cumulative reward by interacting with an environment.
Assume we are given an MDP $(\mathcal S,\mathcal A,P,r,y)$, where $\mathcal S$ is the set of states, $\mathcal A$ describes the set of actions the agent can take, $P:\mathcal S\times \mathcal A\times\mathcal S\to[0,1]$ describes the probability of a transition, $r:\mathcal S\times \mathcal A\to \mathbb R$ rewards the transition and $\gamma\in(0,1)$ is a discount factor.
Setting $\vec a_{t}\sim\pi(\vec s_t)$, $\vec s_{t+1}\sim P(\vec s_t,\vec a_t)$, the goal is
\begin{align}
	\max_{\pi}\max_{\vec a}Q^{\pi}(\vec s_0,\vec a),\quad Q^{\pi}(\vec s_0,\vec a_0)=\E{r(\vec{s}_0,\vec a_0)+\sum_{t=1}^{\infty}\gamma^t r(\vec s_t,\vec a_t)}\;.
\end{align}
In value-based RL, the goal is to learn an accurate approximation of the action-value function $Q$, which in turn yields the optimal policy through the recursive Bellman optimality criterion~\cite{bellman1954theory}.
While deep NNs are often used for approximating $Q$ directly, 
free energy-based RL~(FERL) approximates $Q$ by the free energy of a SQBM.
Partial derivatives are analytically computable $\partial_xF(\vec v)=\tr\left[\rho_{\vec v}\partial_x\mat H\right]$ (see~\cref{sec:fe_gradient_proof}), which corresponds to computing the expectation value of $\partial_x \mat H$ w.r.t. the Gibbs state $\rho_{\vec v}$.
%\begin{align}
%	\partial_w F(\vec v)=-\frac{\tr\left[\partial_w\left(e^{-\beta \mat H}\mat{\Delta}_{\vec v}\right)\right]}{\beta\tr\left[e^{-\beta \mat H}\mat{\Delta}_{\vec v}\right]}
%	=\frac{\tr\left[\left(\partial_w\mat H\right)e^{-\beta \mat H}\mat{\Delta}_{\vec v}\right]}{\tr\left[e^{-\beta \mat H}\mat{\Delta}_{\vec v}\right]}=\tr\left[\rho_{\vec v}\partial_w\mat H\right],
%\end{align}
%with the marginal distribution of the visible units given by $\rho_{\vec v}=e^{-\beta \mat H}\mat{\Delta}_{\vec v}$.
Since $\partial_x\mat H$ can be easily computed due to the quadratic form in~\cref{eq:qbm_hamiltonian}, we obtain the $Q$-learning-based weight update formula with $Q(\vec s,\vec a)=-F(\vec v)$
\begin{align*}
	w\gets w+\alpha \left[\left(F(\vec s,\vec a)+r(\vec{s},\vec a)-\gamma \min_{\vec a'}F(\vec s',\vec a')\right)\partial_wF(\vec s,\vec a)\right]
	\;,
\end{align*}
by encoding $(\vec s,\vec a)$ into the visible units (see~\cref{fig:overview}).
Simply assuming continuous-valued $\vec v$ violates the theoretical foundations of the model, as encoding continuous variables into a quantum state and computing the corresponding marginal distribution is highly non-trivial.
In the following sections, we introduce a principled solution to overcome these challenges.

\section{Continuous Semi Quantum Boltzmann Machines}
\label{sec:continuous}

Instead of considering a joint quantum state over visible and hidden units, we consider a hybrid quantum-classical model.
We assume a prior on the visible units from the exponential family, that is $e^{c(\vec v)-A(\vec\theta)}$, with $c(\vec v)=\vec \theta^{\top}\vec s(\vec v)+\log g(\vec v)$ and $A(\vec\theta)=\log \int_{\vec v}e^{c(\vec v)}\, d\vec v$, similar to~\cite{li2018exponential}.
This leads to the following Hamiltonian of our proposed continuous SQBM~(CSQBM)
\begin{align}
	\mat H(\vec v)=-c(\vec v)\mat I+\mat H^{vh}(\vec v)+\mat H^{h}+\mat H^{hh},
	\quad \mat H^{vh}(\vec v) =-\sum_{ij}\sum_{\mat P} w_{ij}^{vh,\mat P}s_i(\vec v)\mat P_j\;,
	\label{eq:csqbm_hamiltonian}
\end{align}
% \in\mathbb C^{2^m\times 2^m}
where the first summand replaces $\mat H^{v}+\mat H^{vv}$ and the second one replaces $\mat H^{vh}$ in~\cref{eq:qbm_hamiltonian}.
Note that since we do not encode the visible units into qubits anymore, our quantum state space size gets reduced from $\mathbb C^{n+m}$ to $\mathbb C^{m}$.
We obtain a conveniently computable form of the free energy.
\begin{theorem}\label{theo:fe_theorem}
	With $\mat H'(\vec v)=\mat H^{v h}(\vec v)+\mat H^{h}+\mat H^{hh}$ and  $\rho'_{\vec v}=e^{-\beta\mat H'(\vec v)}/\tr\left[e^{-\beta\mat H'(\vec v)}\right]$, it holds
	\begin{align}
		F_{\mat H}(\vec v)=-c(\vec v)+F_{\mat H'}(\vec v),\quad 
		\partial_xF_{\mat H}(\vec v)=-\partial_xc(\vec v)+\tr\left[\rho'_{\vec v}\partial_x\mat H'\right]\;.
		\label{eq:fe_theorem}
	\end{align}
\end{theorem}
A detailed proof is provided in~\cref{sec:fe_theorem_proof}.
Interestingly, we cannot only differentiate over the models' parameters in~\cref{eq:fe_theorem} but also over the visible units.
This leads to the applicability in AC-algorithms, due to the need for differentiating w.r.t. the input action, while overcoming the limitations of the method presented in~\cite{schenk2024hybrid}, which relies on finite differences for computing the gradient and presents a general continuous-variable alternative to the previously imposed binary Bernoulli prior. 

While efficient preparation of Gibbs state is a largely open question, the authors of~\cite{jerbi2021quantum} show that this can be done in a runtime being proportional to smallest spectral gap of a quantum Markov chain. 
When the gap is not exponentially small, the free energy is efficiently estimated in a runtime linear in the complexity of preparing the Gibbs state.

As an example, consider an independent Gaussian prior for every visible unit, that is $v_i\sim\mathcal N(\mu_i,\sigma_i)$.
%\begin{align*}
%	e^{\vec \theta^{\top}\vec s(\vec v)+\log g(\vec v)-A(\vec\theta)}=\prod_i\frac{1}{\sigma_i\sqrt{2\pi}}e^{-\frac{v_i^2}{2\sigma_i^2}+\frac{v_i\mu_i}{\sigma_i^2}-\frac{\mu_i^2}{2\sigma_i^2}}=\prod_i\mathcal N(\mu_i,\sigma_i)\;.
%\end{align*}
The form $e^{c(\vec v)-A(\vec\theta)}$ is obtained by using $\vec s(\vec v)^{\top}=\left(\vec s_1(v_1),\dots,\vec s_n(v_n)\right)$ and $\vec \theta^{\top}=\left(\vec \theta_1,\dots,\vec \theta_n\right)$
\begin{align*}
	\vec s_i(v_i)=\left(v_i,v_i^2\right), 
	\quad \vec \theta_i=\left(\mu_i/\sigma_i^2,-1/2\sigma_i^2\right),
	\quad g(\vec v)=1\quad \Rightarrow c(\vec v)=(-v_i^2+2v_i\mu_i)/2\sigma_i^2\;.
\end{align*}

\section{Continuous Q-Learning}
\label{sec:continuous_q_learning}

Policy-based methods, such as AC algorithms, suffer from critical drawbacks, most notably sample inefficiency.
In contrast, relying solely on the $Q$-learning update for value approximation results in a non-convex global optimization problem.
Our approach is to replace the maximization step in $Q$-learning with sampling, motivated by the observation that $\argmax_{\vec a}Q(\vec s,\vec a)=\argmax_{\vec a}e^{-\beta F(\vec s,\vec a)}=\argmax_{\vec a}p(\vec a|\vec s)$.
While direct sampling from $p(\vec a|\vec s)$ remains infeasible, we instead consider sampling from the marginal distribution of the visible units.
\begin{theorem}\label{theo:marginal}
	Given a configuration $\vec h$ of $\mat H(\vec v)$ w.r.t. to the Pauli measurement basis $\mat P$, the marginal distribution of the visible units is also from the exponential family
	\begin{align}
		p(\vec v|\vec h)=e^{c'(\vec v)-A'(\vec \theta')},
		\quad c'(\vec v)=\vec \theta'^{\top}\vec s(\vec v)+\log g'(\vec v)\;,
	\end{align}
	with $\vec \theta'= \beta\left(\vec\theta+\mat W\vec h\right)$, $W_{ij}=w_{ij}^{vh,\mat P}$,  $g'(\vec v)=g(\vec v)^{\beta}$ and $A'(\vec\theta')=\int_{\vec v}e^{\vec \theta'^{\top}\vec s(\vec v)+\log g'(\vec v)}\, d\vec v$.
\end{theorem}
The proof is given in~\cref{sec:marginal_theorem_proof}.
If the prior is efficiently samplable, so is $p(\vec a|\vec s,\vec h)$.
With the further assumption of an efficiently preparable
Gibbs state of $\mat H'$, sampling from $p(\vec h|\vec s,\vec a)$ is also efficient.
This leads to the applicability of Gibbs sampling by alternatingly generating samples from $p(\vec a|\vec s,\vec h)$ and $p(\vec h|\vec s,\vec a)$ to obtain a sample $\vec a\sim p(\vec a|\vec s)$.
The probability of obtaining the best actions increases with the inverse temperature $\beta$.
Further, samples from $p(\vec a|\vec s)$ can be used for exploration, instead of relying on $\epsilon$-greedy strategies.

\section{Conclusion}
\label{sec:conclusion}

In this work, we introduced Continuous Semi Quantum Boltzmann Machines~(CSQBMs) as a theoretically grounded framework that allows for continuous-action $Q$-learning.
By leveraging exponential family priors for continuous variables and hybrid quantum-classical sampling for action selection, our approach overcomes key limitations of existing Actor-Critic methods, achieving greater expressiveness and sample efficiency while reducing qubit requirements.

For future work, the performance of our methods will be investigated on real-world continuous-control problems---such as particle beam lines at CERN.
Further it is interesting to examine the structure and expressiveness of the underlying Hamiltonians of efficiently preparable Gibbs states.  

\printbibliography

%%%%%%%%%%%%%%%%%%%%%%%%%%%%%%%%%%%%%%%%%%%%%%%%%%%%%%%%%%%%

\newpage
\appendix
	
\section{Technical Appendices and Supplementary Material}

\subsection{Gradient of Free Energy}
\label{sec:fe_gradient_proof}

The free energy of an SQBM Hamiltonian $\mat H$ of the form given in~\cref{eq:qbm_hamiltonian} is defined as 
\begin{align*}
	F_{\mat H}(\vec v)=-\frac{1}{\beta}\log \tr\left[e^{-\beta \mat H}\mat \Delta_{\vec v}\right]
\end{align*}
Due to $\mat He^{-\beta \mat H}=e^{-\beta \mat H}\mat H$ and $\mat H\mat \Delta_{\vec v}=\mat \Delta_{\vec v}\mat H$, we obtain $\partial_x e^{-\beta \mat H}=-\beta e^{-\beta \mat H}\partial_x \mat H$ and thus 
\begin{align*}
	\partial_x F_{\mat H}(\vec v)=-\frac{1}{\beta}\frac{\partial_x \tr\left[e^{-\beta \mat H}\mat \Delta_{\vec v}\right]}{\tr\left[e^{-\beta \mat H}\mat \Delta_{\vec v}\right]}
	=\frac{\tr\left[e^{-\beta \mat H}\mat \Delta_{\vec v}\partial_x \mat H\right]}{\tr\left[e^{-\beta \mat H}\mat \Delta_{\vec v}\right]}
	=\tr\left[\rho_{\vec v}\partial_x\mat H\right]\;.
\end{align*}

\subsection{Proof of \Cref{theo:fe_theorem}}
\label{sec:fe_theorem_proof}

Now assume that $\mat H(\vec v)$ describes the energy of a CSQBM given in~\cref{eq:csqbm_hamiltonian}.
\Cref{theo:fe_theorem} is obtained by
\begin{align*}
	\partial_x F_{H(\vec v)}(\vec v)
	&=-\frac{1}{\beta}\frac{\partial_x \tr\left[e^{-\beta \mat H(\vec v)}\right]}{\tr\left[e^{-\beta \mat H(\vec v)}\right]}
	=-\frac{1}{\beta}\frac{\partial_x \tr\left[e^{-\beta\left(-c(\vec v)\mat I+\mat H'(\vec v)\right)}\right]}{\tr\left[e^{-\beta \left(-c(\vec v)\mat I+\mat H'(\vec v)\right)}\right]} \\
	&=-\frac{1}{\beta}\frac{\partial_x\left(e^{\beta c(\vec v)} \tr\left[e^{-\beta \mat H'(\vec v)}\right]\right)}{e^{\beta c(\vec v)} \tr\left[e^{-\beta \mat H'(\vec v)}\right]} \\
	&=-\frac{1}{\beta}\frac{\partial_xe^{\beta c(\vec v)} \tr\left[e^{-\beta \mat H'(\vec v)}\right]+ e^{\beta c(\vec v)} \partial_x\tr\left[e^{-\beta \mat H'(\vec v)}\right]}{e^{\beta c(\vec v)} \tr\left[e^{-\beta \mat H'(\vec v)}\right]} \\
	&= \frac{-\partial_xc(\vec v) \tr\left[e^{-\beta \mat H'(\vec v)}\right]+ \tr\left[e^{-\beta \mat H'(\vec v)}\partial_x\mat H'(\vec v)\right]}{ \tr\left[e^{-\beta \mat H'(\vec v)}\right] }
	=- \partial_x c(\vec v)+\tr\left[\rho'_{\vec v}\partial_x\mat H'(\vec v)\right]\;,
\end{align*}
where we used the fact that $e^{\beta \mat I+\mat H}=e^{\beta \mat I}e^{\mat H}=e^{\beta}e^{\mat H}$ for the identity matrix $\mat I$.

\subsection{Proof of \Cref{theo:marginal}}
\label{sec:marginal_theorem_proof}

Assume we are given a hidden configuration $\vec h\in\{-1,1\}^m$ w.r.t. to some Pauli basis $\mat P$.
Computing the trace of a Hamiltonian only acting on the hidden units $\mat H^{h}+\mat H^{hh}$, leads to a scalar $\lambda(\vec h)$ independent of $\vec v$.
Thus
\begin{align*}
	p(\vec v|\vec h)
	&=\frac{\tr\left[e^{-\beta\mat H(\vec v)}\mat \Delta_{\vec h}\right]}{\int_{\vec v}\tr\left[e^{-\beta\mat H(\vec v)}\mat \Delta_{\vec h}\right]\,d\vec v}
	=\frac{e^{-\beta\left(-c(\vec v)-(\mat W\vec h)^{\top}\vec s(\vec v)+\lambda(\vec h)\right)}}{\int_{\vec v}e^{-\beta\left(-c(\vec v)-(\mat W\vec h)^{\top}\vec s(\vec v)+\lambda(\vec h)\right)}\,d\vec v}  \\
	&=\frac{
		e^{\beta\left(
			\vec{\theta}^{\top}\vec s(\vec v)
			+\log g(\vec v)
			+(\mat W\vec h)^{\top}\vec s(\vec v)
			\right)}
	}
	{\int_{\vec v}e^{\beta\left(\vec\theta^{\top}\vec s(\vec v)+\log g(\vec v)+(\mat W\vec h)^{\top}\vec s(\vec v)\right)}\,d\vec v} \\
	&=\frac{
		e^{c'(\vec v)}
	}
	{\int_{\vec v}e^{c'(\vec v)}\,d\vec v}
	=e^{c'(\vec v)-A'(\vec \theta')}\;,
\end{align*}
with $c'(\vec v)=\vec \theta'^{\top}\vec s(\vec v)+\log g(\vec v)^{\beta}$, $\vec \theta'=\beta(\vec \theta+\mat W\vec h)$ and $W_{ij}=w_{ij}^{vh,\mat P}$.

\end{document}